%% file: paper_arxiv.tex
\documentclass{article}
\usepackage{iclr2027_conference,times}
\input{math_commands.tex}

\usepackage{amsmath,amssymb,booktabs,multirow,microtype}
\usepackage{algorithm}
\usepackage{algpseudocode}
\usepackage{graphicx}
\usepackage{tikz}
\usetikzlibrary{arrows.meta,positioning,fit}
\usepackage[hidelinks]{hyperref}
\usepackage{url}

\title{Semantic Context Handoff with On-Demand Evidence for Long-Horizon Agents}

\author{
Guolong Zheng, Chen Hou, Xu Yang, Jianshan Zhang, Tao Huang \\
Minjiang University \\
Fuzhou, China \\
\texttt{\{gzheng, houchen, xu.yang, jszhang, huang-tao\}@mju.edu.cn}
}

\iclrfinalcopy

\begin{document}
\maketitle
\begin{abstract}
Traditional long-horizon LLM agents often hand off context through full-history prompting, retaining all previous action--observation interactions. As execution proceeds, this history keeps growing, increasing inference cost and potentially degrading accuracy. Existing methods control this growth by preserving compact conclusions while removing earlier observations, but doing so can discard evidence that later becomes relevant. We introduce \textbf{Semantic Context Handoff (SCH)}, whose central abstraction is \emph{state--evidence separation}: a compact semantic state carries the agent's current interpretation, while supporting action--observation evidence remains separately addressable and can re-enter context when later execution makes it relevant, without separate summarization or retrieval-model calls.

Empirically, SCH improves long-horizon agent performance while substantially reducing context cost. In a controlled 150-task SWE-bench study, SCH's evidence recovery improves Pass@1 from $80.0\%$ to $86.9\%$ at essentially unchanged token use and execution time. In the complete-system comparison, SCH achieves $86.9\%$ Pass@1 while using $43\%$ fewer tokens and $41\%$ less wall-clock time than the strongest baseline by task performance. Across models and benchmarks, SCH consistently reduces token use relative to full-history prompting while remaining competitive with or improving upon the evaluated baselines. Trajectory analysis confirms that SCH recovers evidence from substantially earlier parts of the interaction history rather than merely preserving a recent context window.
\end{abstract}

\section{Introduction}
\label{sec:introduction-main}
Long-horizon agents repeatedly interact with an environment and carry the resulting information into subsequent prompts. Retaining every action and observation preserves the complete execution history, but repeatedly resubmitting an ever-growing context is expensive and can bury currently relevant information \citep{liu2024lost}. Compressing or summarizing the context controls this growth, yet creates a different failure mode: a compact conclusion may remain in context after the observations needed to reconsider that conclusion have been removed \citep{wu2025resum,kang2026acon,huang2026pact}.

The usefulness of historical evidence is not determined solely by chronological order. A software-engineering agent may inspect a file, form a root-cause hypothesis, modify the repository, and many steps later encounter a failed verification that makes the original inspection relevant again. A recent-window strategy may already have discarded that observation, while an irreversible summary may preserve only the resulting hypothesis. This creates two retention requirements: the agent's current understanding should remain compact and continuously available, while detailed historical observations need only remain recoverable when later execution makes them relevant again. Full-history prompting preserves both at growing context cost, whereas irreversible compression can sacrifice recoverability. We therefore separate two questions: \emph{what semantic information should remain continuously visible, and what historical evidence should remain recoverable when it becomes relevant again?}

We introduce \textbf{Semantic Context Handoff (SCH)} to address these requirements. At each interaction step, SCH hands the LLM a compact context containing a \emph{semantic state}---its current understanding and work status---together with \emph{evidence} relevant to the current decision. The semantic state captures task constraints, revisable facts and hypotheses, and work lifecycle, while \emph{evidence} consists of stored action--observation records that support the current state or may become relevant to later decisions. These records are preserved separately in an evidence archive, allowing earlier observations to leave the routine prompt and later return when subsequent execution signals that they are relevant again. Rather than regenerating the full state after each interaction, the model proposes only a sparse state delta alongside its next action, while a deterministic controller updates the state and selects evidence for subsequent handoffs.

Our experiments test both the recovery mechanism and the resulting complete system. In a controlled ablation with semantic-state machinery, controller policy, action interface, and evidence budget fixed, recovering evidence associated with previously completed work increases mean Pass@1 from $80.00\%$ to $86.89\%$ at essentially unchanged token use and execution time, with significant paired improvements across all three repeated runs. In the complete-system comparison, SCH reaches $86.89\%$ mean Pass@1 while using $43.3\%$ fewer tokens and $40.8\%$ less wall-clock time than the strongest baseline by task performance. Across models and benchmarks, SCH attains the highest observed Pass@1 in three of four conditions while providing a strong effectiveness--efficiency trade-off. Trajectory analysis further confirms that SCH recovers evidence from substantially earlier parts of the interaction history rather than merely preserving a recent context window.

The key contribution is a different representation boundary for long-horizon context management: \emph{state--evidence separation}. SCH separates the agent's mutable interpretation of execution from the append-only action--observation evidence on which it was formed. The former is maintained as compact semantic state; the latter may leave routine context while remaining addressable through provenance. Lifecycle-aware revision reconnects them when later execution invalidates a conclusion or reopens completed work. Structured state and external memory have precedents \citep{kim2025coex}; SCH's contribution is this boundary together with revision semantics that operate across it.

Concretely, we make the following contributions:
\begin{itemize}
\setlength{\itemsep}{1pt}
\setlength{\parskip}{0pt}
\item \textbf{State--evidence separation for semantic handoff.} SCH keeps mutable current understanding in compact semantic state and executed action--observation evidence in a separate append-only archive. Provenance links state entries to the archive, so compact context need not losslessly preserve its source evidence.

\item \textbf{Revision-aware evidence recovery.} Evidence associated with completed work can leave routine context and later become available again when execution indicates that the corresponding state should be reconsidered, without an additional memory or retrieval-model call.

\item \textbf{Controlled and broad evaluation.} Matched ablations isolate the contribution of historical evidence recovery, while complete-system, cross-model, cross-benchmark, component, and trajectory analyses evaluate effectiveness, efficiency, portability, and non-local evidence recovery.
\end{itemize}

Appendix~\ref{sec:introduction} provides additional motivation, and Appendix~\ref{sec:related-work} provides expanded related-work distinctions omitted from the main paper for space.

\section{Related Work}
\label{sec:related-main}

\paragraph{Compression and context management.}
Long-context methods reduce active history through several forms of context compression. LongLLMLingua \citep{jiang2024longllmlingua} compresses prompts by selectively retaining informative content, while ReSum \citep{wu2025resum} periodically summarizes accumulated interaction history and ACON \citep{kang2026acon} learns compression strategies for long-horizon agents. PACT \citep{huang2026pact} represents interactions as compact action--state records, whereas CAT \citep{liu2026context} and MemPO \citep{li2026mempo} learn when and how context or memory should be condensed and retained. SCH differs in what it carries forward: rather than maintaining a compressed interaction history, it maintains a single representation of the current execution state---task constraints, revisable facts and hypotheses, and work lifecycle---and updates it through sparse deltas. Detailed action--observation evidence is stored separately, allowing the routine handoff to remain centered on the agent's current task state while historical details remain recoverable when needed.

\paragraph{Agent memory and recoverable information.}
A broader line of work stores, abstracts, retrieves, or reuses prior experience to support later reasoning. Generative Agents \citep{park2023generative} and MemoryBank \citep{zhong2024memorybank} maintain long-term experience or conversational memory. Reflexion \citep{shinn2023reflexion} and ExpeL \citep{zhao2024expel} retain feedback or reusable experience from previous attempts, while Agent Workflow Memory \citep{wang2025awm} and Voyager \citep{wang2023voyager} reuse learned workflows or skills across tasks. MemGPT \citep{packer2023memgpt} organizes information across memory tiers, whereas ReadAgent \citep{lee2024readagent} and OCR-Memory \citep{li2026ocrmemory} preserve access to higher-fidelity source content after compaction. These methods establish the value of retaining and selectively recovering information rather than continuously exposing the full history. SCH focuses specifically on linking a compact representation of the current execution state to the underlying action--observation evidence that may later be needed to reconsider it.

\paragraph{Structured and revision-aware memory.}
Recent work also moves beyond free-form summaries toward explicit state representations. HiSA \citep{li2026hisa} hierarchically transforms interaction history into compact semantic states, while CoEx \citep{kim2025coex} maintains an evolving neurosymbolic belief state. SCH therefore does not claim that structured or semantic state itself is new. The distinction is what the state is responsible for preserving: SCH deliberately makes the semantic state non-self-sufficient, because observations needed to verify or revise it remain separately addressable. CoEx evolves the belief state; MAGE \citep{chen2026mage} can restore an earlier execution-state boundary; and Proactive Memory Agent \citep{wu2026proactivememory} uses a separate memory agent to decide what stored information to surface. SCH instead keeps current interpretation and source evidence as distinct persistent objects and defines revision as selective re-exposure of completed-work provenance. Additional distinctions are provided in Appendix~\ref{sec:related-work}.

\section{Semantic Context Handoff}
\label{sec:method-main}

\begin{figure}[tb]
\centering
\includegraphics[width=0.72\linewidth, trim=6 4 6 4, clip]{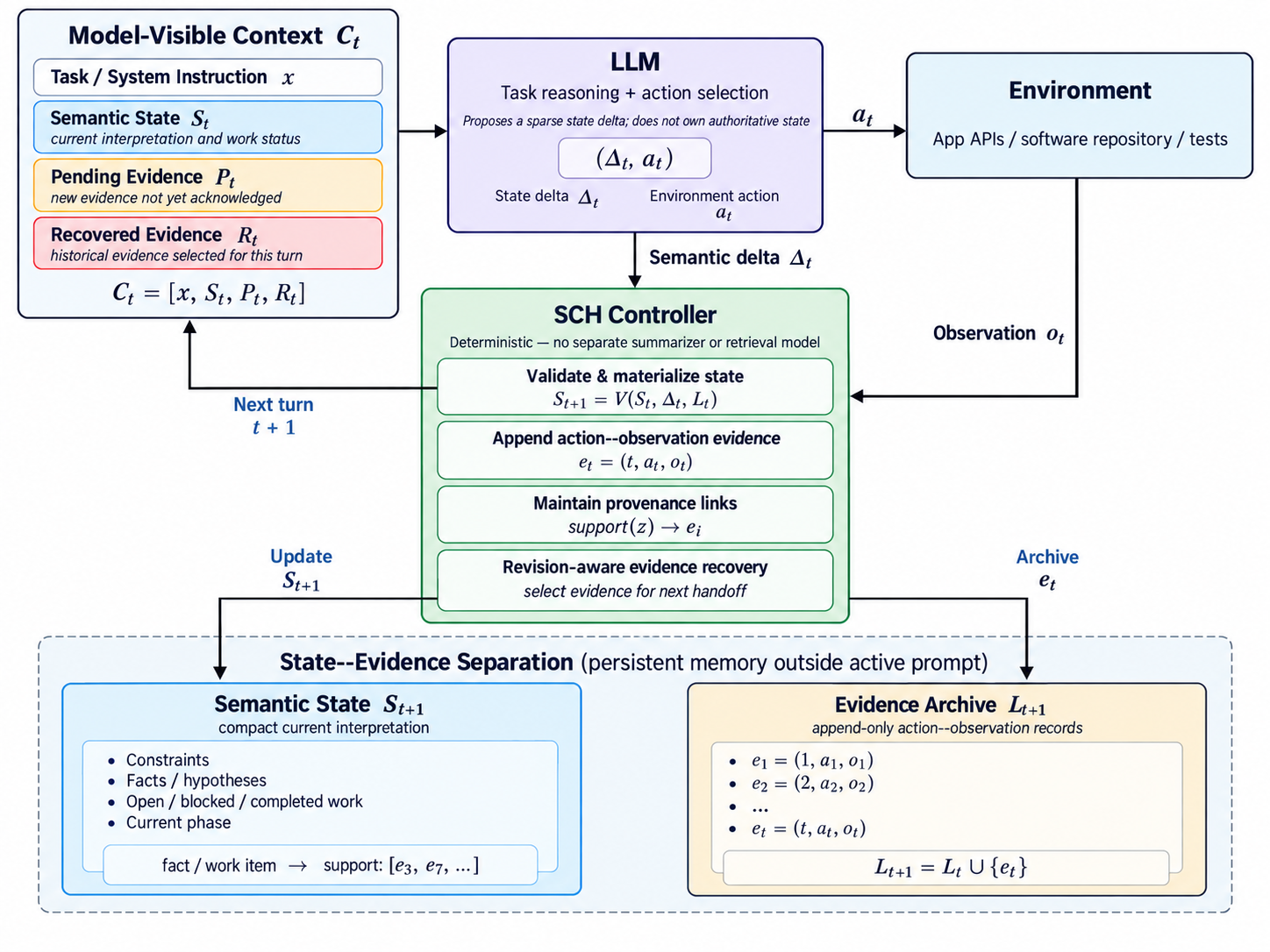}
\caption{\textbf{SCH workflow.} The LLM receives the current semantic state together with relevant evidence and returns a sparse state delta and environment action. The controller updates the state, executes the action, stores the resulting action--observation evidence, and selects evidence for subsequent handoffs.}
\label{fig:sch-workflow-main}
\end{figure}

As shown in Figure~\ref{fig:sch-workflow-main}, SCH organizes each agent interaction around three components: a \emph{semantic state}, an \emph{evidence archive}, and a deterministic \emph{controller}. The semantic state compactly represents the agent's current understanding and work status, while the evidence archive preserves executed action--observation records outside the routine prompt. The controller connects the two by updating the state and determining which archived evidence should become visible again.

Each interaction follows the same handoff loop. At step $t$, the controller constructs the LLM handoff prompt

$$
C_t=[x,\;S_t,\;P_t,\;R_t],
$$

where $x$ is the task/system instruction, $S_t$ is the current semantic state, $P_t$ contains recently produced evidence that remains visible until processed by the LLM, and $R_t\subseteq L_t$ contains historical evidence recovered from evidence archive $L_t$ for the current decision. The LLM returns two outputs in one response: a sparse state delta $\Delta_t$ describing changes to $S_t$, and the next environment action $a_t$. The controller applies the delta to obtain $S_{t+1}$, executes $a_t$, stores the resulting action--observation record in $L_{t+1}$, and constructs the next handoff $C_{t+1}$. The controller is a deterministic, non-LLM component; state reconciliation, evidence storage, and evidence selection therefore require no separate summarization or retrieval-model call.

\subsection{Semantic State and Sparse Deltas}
\label{sec:state-main}

The semantic state $S_t$ represents the agent's \emph{current execution status} rather than a compressed transcript of previous interactions. It contains four types of information: \emph{task constraints}, \emph{facts and hypotheses}, \emph{work items}, and a \emph{task phase}. Task constraints capture requirements or assumptions expected to remain stable. Facts capture environment-grounded information that the agent currently treats as established---for example, that an inspected component has a particular property or that a tool returned a particular result---whereas hypotheses capture provisional interpretations that may later be revised. These labels organize the agent's current state rather than certify semantic truth.

Work items represent responsibilities or subgoals and follow a simple lifecycle: \emph{open} work remains unresolved, \emph{blocked} work cannot currently proceed, and \emph{completed} work is considered resolved. For example, ``identify the cause of failure $F$'' may begin as open, become blocked while required information is unavailable, and later become completed after a diagnosis is established. The task phase gives a coarse description of the current execution stage, such as information gathering, acting, or verification.

Facts, hypotheses, and work items can retain \emph{provenance links} to the evidence on which they were based. Concretely, these links are evidence-record identifiers stored with a state entry. \emph{Support} identifies evidence directly cited for a fact, hypothesis, or work item, while \emph{lineage} preserves evidence accumulated as a work item progresses through its lifecycle. For example, if observation $e_7$ motivates hypothesis $h_1$, the LLM can attach $e_7$ as support for $h_1$. If the corresponding work is later completed, its provenance links remain associated with that work rather than being discarded.

Instead of asking the LLM to reproduce the entire state after every interaction, the prompt requests only a sparse update, \(\{\texttt{"delta"}:\Delta_t,\;\texttt{"command"}:a_t\}\). This preserves unchanged state entries rather than asking the LLM to repeatedly reconstruct them, reducing both redundant generation and unintended modification of state that is not being revised. The delta therefore contains only changes caused by the current interaction, such as adding or revising a fact or hypothesis, changing a work item's status, reopening completed work, or marking evidence as processed.

The controller computes

$$
S_{t+1}=\mathcal{V}(S_t,\Delta_t,L_t),
$$

where $\mathcal{V}$ deterministically reconciles the proposed delta with the previous state. Evidence identifiers referenced by the delta are checked against $L_t$, and requested lifecycle transitions are checked for structural validity. The controller therefore maintains state structure and provenance, while semantic reasoning remains the responsibility of the LLM. Additional state fields and reconciliation rules are provided in Appendix~\ref{sec:semantic-state}--\ref{sec:sparse-handoff}.

\subsection{Evidence Archive and Acknowledgement}
\label{sec:evidence-main}

The evidence archive $L_t$ stores the actions actually executed in the environment and the observations they return. After the environment executes $a_t$ and produces observation $o_t$, the controller forms evidence record $e_t=(t,a_t,o_t)$ and updates the archive by adding $e_t$ to $L_t$ to obtain $L_{t+1}$. The step number $t$ serves as the evidence identifier used by the support and lineage links defined above. Model reasoning is not stored as execution evidence.

A newly produced record remains visible in $P_t$ until the LLM has processed it. The LLM then \emph{acknowledges} the record, indicating either that its durable implication has been incorporated into the semantic state or that the observation does not need persistent visibility. For example, an observation may support completing an open work item, in which case its implication is retained in that work item's state; alternatively, an inspection may prove irrelevant and require no persistent state update. In either case, acknowledgement removes the record from pending context $P_t$ but not from the evidence archive $L_t$.

Thus, acknowledgement changes \emph{visibility}, not storage. Evidence can leave the routine handoff without being deleted and remain available for later recovery.

\subsection{Revision-Aware Evidence Recovery}
\label{sec:recovery-main}

The provenance links defined above provide the primary route from semantic state back to historical evidence. During ordinary execution, the controller can select support and lineage associated with currently retained facts, hypotheses, and unresolved work for inclusion in $R_t$.

A core SCH mechanism handles cases in which previously completed work must be reconsidered. A \emph{revision signal} can arise in two ways. First, the controller can detect a recognized failure in newly produced environment evidence while revisable state is present. Second, the LLM can explicitly revise the state through its delta, for example by retracting a fact or reopening completed work. In either case, provenance associated with the affected completed work becomes eligible for recovery in addition to evidence associated with the currently active state. This allows observations from much earlier execution stages to return even after the work they supported has left the active handoff.

A concrete example illustrates the mechanism. Suppose an agent inspects a system at step $6$ and obtains evidence $e_6$ indicating that configuration $A$ may cause failure $F$. It records hypothesis $h_1$ with $e_6$ as support, modifies the system, and later marks the corresponding work completed. After acknowledgement, $e_6$ no longer needs to remain in the routine prompt. At step $24$, however, verification produces a new failure. The controller recognizes the failure while revisable state exists, or the LLM responds by reopening the completed work in its delta. Either event triggers revision, making the completed work's provenance---including $e_6$---eligible for the next handoff. The LLM can then reconsider the earlier diagnosis using the original observation rather than only its compact state representation.

Provenance cannot cover every observation that may later become useful. Some evidence may never have been attached to a persistent state entry, or its previous state entry may no longer be retained. SCH therefore supplements provenance with deterministic \emph{orphan recovery}: archived records without a current provenance link can become candidates when their paths or terms overlap with the current state. Recent and safety-relevant records provide additional fallback candidates. These sources complement rather than replace provenance: directly linked evidence receives highest priority, while revision additionally activates completed-work provenance. Algorithm~\ref{alg:recovery-main} summarizes the resulting recovery procedure.

\begin{algorithm}[tbh]
\caption{SCH historical evidence recovery}
\label{alg:recovery-main}
\begin{algorithmic}[1]
\Require semantic state $S$, evidence archive $L$, pending evidence $P$
\State collect support and lineage linked to currently relevant state entries
\State detect revision from recognized failure evidence or an accepted state delta
\If{revision is triggered}
\State add provenance associated with affected completed work
\EndIf
\State add deterministic orphan, recent, and safety-relevant candidates
\State merge and rank candidates, excluding evidence already visible in $P$
\State expose the highest-priority evidence under the configured handoff budget
\end{algorithmic}
\end{algorithm}

Exact orphan-selection rules, ranking details, candidate quotas, and rendering limits are implementation parameters described in Appendix~\ref{sec:evidence-recovery}. Section~\ref{sec:experiments-main} directly evaluates the key distinction between recovery from currently active state and full SCH recovery that additionally makes completed-work provenance available under revision.

\paragraph{Putting the pieces together.}
Consider a generic task in which an agent must modify a system to satisfy requirement $R$ while preserving constraint $C$. After several interactions, a handoff might contain:
\begin{quote}
\small
\texttt{TASK:} satisfy requirement $R$ while preserving $C$.\
\texttt{SEMANTIC STATE:}\
\quad constraint: preserve $C$;\
\quad fact: the target currently uses option $A$;\
\quad hypothesis: option $A$ may cause failure $F$ [support: $e_4$];\
\quad work: determine whether $A$ causes $F$ [open].\
\texttt{RELEVANT EVIDENCE:}\
\quad $e_7$: action = verify target; observation = failure $F$ observed.\
\texttt{OUTPUT:} sparse state delta + next environment action.
\end{quote}
The archive may contain many other action--observation records, but only evidence selected for the current decision appears in the handoff. Here, the semantic state carries the agent's current understanding, $e_7$ provides evidence relevant to the immediate decision, and the sparse delta allows the LLM to update only the affected state entries. If later execution invalidates a conclusion or reopens completed work, the lifecycle-aware mechanism described above can bring its earlier supporting evidence back into a subsequent handoff.

\section{Experiments}
\label{sec:experiments-main}

\paragraph{Setup.}
We evaluate SCH on AppWorld \citep{trivedi2024appworld} and SWE-bench Verified \citep{jimenez2024swebench} with MiniMax and DeepSeek. Because repeating every method--model--environment combination is computationally prohibitive, we use three provider runs in one primary setting for controlled mechanism attribution and frozen single-run breadth settings to test transfer. SWE-bench uses a 150-task sample from the 450 tasks remaining after a separate 50-task development cohort. For stratification only, we compute a deterministic difficulty proxy from gold-patch changed lines, files, and hunks, divide candidates into terciles, and sample jointly by repository and difficulty, yielding 49 easy, 50 medium, and 51 hard tasks shared by all SWE-bench methods and both model families. AppWorld uses the 417-task test-challenge cohort for MiniMax and a separate frozen random 200-task sample for DeepSeek, so its model blocks are not a controlled cross-model comparison. AppWorld and SWE-bench also use different host-agent interfaces (the AppWorld agent interface and SWE-Agent \citep{yang2024sweagent}); SCH keeps the same state--evidence architecture and controller logic across them, with adaptation limited to environment-facing prompt/action formatting and execution-event handling. Exact task lists, sampling metadata, and run manifests are in the anonymous repository.

DeepSeek SWE-bench is the \emph{primary setting}. Full History, PACT, ACON, SCH, and controlled evidence selectors each receive three runs on the same 150 tasks; breadth settings are single-run. PACT/ACON are matched single-agent adaptations. The primary setting also includes MAGE matched, a paper-guided implementation because sufficient public reference code was unavailable, and Proactive Memory, adapted from its official implementation.

Primary runs use temperature $0$ and top-$p=1$. SWE-bench controller limits are 12,000 characters each for state and recovered evidence, 12 historical records per handoff, candidate caps of six revision-associated/three recent/four orphan/six safety records, a 12-inspection anti-overresearch threshold, 16,384 output tokens, and 100 calls/task. These limits are fixed across controlled selector variants; further details are in Appendix~\ref{sec:experiments} and the anonymous repository.

\paragraph{Evaluation logic and metrics.}
The design provides three complementary forms of evidence: repeated DeepSeek SWE-bench runs isolate the recovery mechanism, component variants diagnose the complementary roles of state and evidence, and the remaining model--environment settings test transfer; the trajectory audit then asks whether recovered evidence is actually non-local. We report Pass@1, cumulative model Tokens/task, and end-to-end wall-clock Time/task. Primary results are mean $\pm$ sample SD over three provider runs. Mechanism-audit statistics use the primary run; active-state provenance versus SCH receives exact paired McNemar tests in all three runs, while other paired task-level and bootstrap analyses use the primary run unless stated otherwise. The broader component analysis is in Appendix~\ref{sec:component-analysis}; Semantic State Only comes from the earlier 50-task development cohort and is diagnostic rather than a matched recovery estimate.

\subsection{Controlled Mechanism Study: Revision-Aware Recovery}
\label{sec:mechanism-main}

The central mechanistic claim of SCH is that evidence associated with previously completed work should become available again when later execution requires the current state to be reconsidered. We therefore compare historical-evidence selection policies while holding the semantic-state machinery, work lifecycle, task cohort, model, action interface, controller limits, and maximum evidence budget fixed. In particular, \emph{active-state provenance} can recover support and lineage associated with currently active state, whereas revision-aware SCH additionally makes provenance from completed work eligible when revision is triggered.

\begin{table*}[t]
\centering
\caption{\textbf{Controlled mechanism study} on frozen 150-task DeepSeek SWE-bench. State/lifecycle machinery, action interface, controller limits, and evidence budget are fixed; only historical evidence selection changes. Mean $\pm$ sample SD over three runs.}
\label{tab:retrieval-main}
\scriptsize
\setlength{\tabcolsep}{5pt}
\begin{tabular}{lrrr}
\toprule
Selection rule & Pass@1 mean $\pm$ SD & Tok./task mean $\pm$ SD & Time/task mean $\pm$ SD \\
\midrule
Recent-window recovery & $80.89\pm2.04\%$ & $169.2\pm4.4$k & $278.7\pm8.0$ s \\
Lexical/BM25 recovery & $83.33\pm1.34\%$ & $182.7\pm4.4$k & $265.2\pm6.2$ s \\
Active-state provenance & $80.00\pm1.33\%$ & $209.8\pm6.2$k & $\mathbf{233.8\pm7.3}$ s \\
\textbf{Revision-aware SCH} & $\mathbf{86.89\pm1.39\%}$ & $\mathbf{208.6\pm5.8}$k & $237.1\pm7.4$ s \\
\bottomrule
\end{tabular}
\end{table*}

Active-state provenance reaches $80.00\pm1.33\%$ mean Pass@1, whereas revision-aware SCH reaches $86.89\pm1.39\%$, a 6.89-point difference at essentially unchanged token use and runtime. Active-state provenance uses $209.8\pm6.2$k tokens/task versus $208.6\pm5.8$k for SCH, with corresponding runtimes of $233.8\pm7.3$ and $237.1\pm7.4$ seconds/task.

The paired task outcomes consistently favor revision-aware SCH across all three runs. SCH-only versus active-state-only outcomes are 15--4 in Run~1, 14--4 in Run~2, and 14--4 in Run~3, with exact McNemar $p=0.0192$, $0.0309$, and $0.0309$, respectively. Thus, under matched state machinery and evidence budgets, making completed-work provenance available under revision produces a consistent and statistically significant paired improvement in all three runs.

The generic selectors provide additional context. Revision-aware SCH exceeds recent-window recovery by 6.00 points and lexical/BM25 recovery by 3.56 points in repeated-run mean Pass@1. Active-state provenance alone does not outperform these generic selectors, so the results do not support a monotonic hierarchy in which provenance is always preferable to recency or lexical retrieval. Instead, they support the narrower mechanism motivating SCH: provenance becomes substantially more useful when revision can reactivate evidence associated with previously completed work. Recent-window and lexical recovery use fewer tokens, while active-state provenance is marginally faster; the direct active-state-versus-SCH comparison should therefore be interpreted primarily as an effectiveness comparison at essentially matched resource cost.

\subsection{Main System Results in the Primary Setting}
\label{sec:primary-results-main}

We next evaluate SCH as a complete context-management system in the repeated DeepSeek SWE-bench setting.

\begin{table*}[t]
\centering
\caption{\textbf{Primary complete-system comparison} on frozen 150-task DeepSeek SWE-bench. Solved/Eval. aggregates three runs on the same tasks; Pass@1 is mean $\pm$ sample SD across runs. MAGE is paper-guided; Proactive Memory is adapted from its official implementation.}
\label{tab:primary-main}
\scriptsize
\setlength{\tabcolsep}{4.2pt}
\begin{tabular}{lrrrr}
\toprule
Method & Solved/Eval. & Pass@1 $\uparrow$ & Tok./task $\downarrow$ & Time/task $\downarrow$ \\
\midrule
Full History & $306/(3\times150)$ & $68.00\pm2.00\%$ & $1713.3\pm73.3$k & $633.6\pm23.8$ s \\
PACT & $377/(3\times150)$ & $83.78\pm1.68\%$ & $367.9\pm11.0$k & $400.2\pm13.8$ s \\
ACON & $364/(3\times150)$ & $80.89\pm1.02\%$ & $520.6\pm15.7$k & $569.6\pm20.7$ s \\
MAGE matched & $230/(3\times150)$ & $51.11\pm2.34\%$ & $389.2\pm23.8$k & $392.8\pm21.3$ s \\
Proactive Memory & $348/(3\times150)$ & $77.33\pm2.00\%$ & $854.4\pm38.3$k & $361.8\pm16.8$ s \\
\textbf{SCH} & $\mathbf{391/(3\times150)}$ & $\mathbf{86.89\pm1.39\%}$ & $\mathbf{208.6\pm5.8}$k & $\mathbf{237.1\pm7.4}$ s \\
\bottomrule
\end{tabular}
\end{table*}

SCH has the highest observed mean Pass@1 in the primary comparison while also using the fewest tokens and the least wall-clock time among the reported methods. PACT is the strongest baseline by observed mean task success. Relative to PACT, SCH uses $208.6\pm5.8$k versus $367.9\pm11.0$k tokens/task and $237.1\pm7.4$ versus $400.2\pm13.8$ seconds/task, corresponding to 43.3\% lower mean token use and 40.8\% lower mean runtime. Run~1 paired task-bootstrap intervals are 31.6--53.0\% for token reduction and 29.0--51.5\% for runtime reduction.

The observed accuracy difference to PACT is more modest. Across the three provider runs, SCH obtains 87.33\%, 85.33\%, and 88.00\% Pass@1, compared with 84.00\%, 82.00\%, and 85.33\% for PACT. In Run~1, 119 tasks are solved by both methods, 12 only by SCH, 7 only by PACT, and 12 by neither, yielding exact McNemar $p=0.359$. We therefore treat the 3.11-point difference in repeated-run means as descriptive rather than as a statistically established paired accuracy advantage.

The primary setting also contains the closest architectural comparisons. SCH reaches $86.89\pm1.39\%$ Pass@1 versus $51.11\pm2.34\%$ for MAGE matched and $77.33\pm2.00\%$ for Proactive Memory. Relative to MAGE matched, SCH uses 46.4\% fewer tokens and 39.6\% less wall-clock time. Relative to Proactive Memory, SCH uses 75.6\% fewer tokens and 34.5\% less wall-clock time. The MAGE result should be interpreted as a controlled paper-guided architectural comparison rather than an official reproduction; the Proactive Memory adaptation follows its released implementation under the common evaluation protocol.

\subsection{Generalization Across Models and Environments}
\label{sec:breadth-main}

We next test whether the complete SCH design remains effective and context-efficient outside the primary DeepSeek SWE-bench setting. Table~\ref{tab:breadth-main} reports the other three model--benchmark conditions; the fourth condition, DeepSeek SWE-bench, is reported in Table~\ref{tab:primary-main}.

\begin{table*}[t]
\centering
\caption{\textbf{Breadth beyond the primary setting.} All rows are single-run. DeepSeek SWE-bench appears in Table~\ref{tab:primary-main}. AppWorld uses 417 MiniMax tasks and a separate frozen 200-task DeepSeek sample, so its blocks are not a controlled cross-model comparison.}
\label{tab:breadth-main}
\scriptsize
\setlength{\tabcolsep}{3.8pt}
\begin{tabular}{lllrrrr}
\toprule
Benchmark & Model & Method & Solved/Eval. & Pass@1 $\uparrow$ & Tok./task $\downarrow$ & Time/task $\downarrow$ \\
\midrule
\multirow{4}{*}{AppWorld (417)} & \multirow{4}{*}{MiniMax}
& Full History & 352/417 & 84.41\% & 341.5k & 182.5 s \\
&& PACT & 296/417 & 70.98\% & \textbf{183.7k} & \textbf{98.6 s} \\
&& ACON & 261/417 & 62.59\% & 300.0k & 175.0 s \\
&& \textbf{SCH} & 360/417 & \textbf{86.33\%} & 223.4k & 114.2 s \\
\midrule
\multirow{4}{*}{AppWorld (200)} & \multirow{4}{*}{DeepSeek}
& Full History & 172/200 & 86.00\% & 522.2k & 374.6 s \\
&& PACT & 174/200 & 87.00\% & \textbf{344.2k} & \textbf{131.7 s} \\
&& ACON & 166/200 & 83.00\% & 497.7k & 300.5 s \\
&& \textbf{SCH} & 177/200 & \textbf{88.50\%} & 408.6k & 247.0 s \\
\midrule
\multirow{4}{*}{SWE-bench (150)} & \multirow{4}{*}{MiniMax}
& Full History & 97/150 & 64.67\% & 928.4k & 385.6 s \\
&& PACT & 113/150 & \textbf{75.33\%} & 723.3k & 368.9 s \\
&& ACON & 100/150 & 66.67\% & 738.3k & 378.9 s \\
&& SCH & 111/150 & 74.00\% & \textbf{491.0k} & \textbf{231.5 s} \\
\bottomrule
\end{tabular}
\end{table*}

SCH has the highest observed Pass@1 in both additional AppWorld settings. On MiniMax SWE-bench, PACT resolves 113/150 tasks compared with 111/150 for SCH, a 1.33-point difference, while SCH uses 491.0k rather than 723.3k tokens/task and 231.5 rather than 368.9 seconds/task. Relative to Full History, SCH reduces token use in every additional condition; on MiniMax AppWorld, for example, it resolves 360/417 tasks while reducing tokens/task by 34.6\% and wall-clock runtime by 37.4\%.

Together, the experiments provide complementary evidence. The repeated DeepSeek SWE-bench study supplies the controlled setting for mechanism attribution, while the remaining frozen model--benchmark conditions test transfer of the complete architecture. SCH has the highest observed Pass@1 in three of four conditions and reduces token use relative to Full History in all four. The same core SCH architecture is instantiated across the AppWorld agent interface and SWE-Agent despite their different interaction interfaces. We therefore interpret the breadth results as cross-model and cross-environment transfer, while reserving causal attribution of revision-aware recovery for the repeated controlled study.

Trajectory inspection suggests that MiniMax's lower SWE-bench result partly reflects more invalid state deltas, preventing some updates; this is less prominent on AppWorld. We therefore treat cross-model differences as partly interface-sensitive rather than purely capability differences.

\subsection{Mechanism Use: SCH Recovers Non-Local Evidence}
\label{sec:mechanism-use-main}

We audit primary-run trajectories to test whether the proposed recovery pathways are exercised. Of 150 SCH trajectories, 83 contain an explicit revision event; direct-support, lineage, orphan, and recent-causal recovery occur in 134, 136, 70, and 116 tasks. Recovered old evidence is 18.48 interaction steps old on average; 95.6\% of occurrences are more than three steps old and 65.9\% more than ten steps old. The 12-record historical-evidence cap is active in 65/150 trajectories.

Thus SCH routinely surfaces non-local evidence rather than merely reproducing a recent interaction tail. Appendix~\ref{app:sympy-case} reconstructs a complete revision episode; Appendix~\ref{sec:experiments} reports the full audit and context-budget diagnostics.

\section{Discussion}
\label{sec:discussion-main}
The selector, component, and complete-system studies answer complementary questions. Semantic State Only is deliberately incomplete: its weaker result supports the \emph{state--evidence separation} premise rather than arguing against semantic state. SCH allows state to be lossy with respect to historical detail because underlying execution evidence remains recoverable. Recent-history variants retain local evidence but lack durable state/provenance links for reactivating completed-work evidence. The matched selector study isolates this distinction: revision-aware recovery raises Pass@1 from $80.00\pm1.33\%$ to $86.89\pm1.39\%$ over active-state provenance at essentially matched token use. Thus state and recoverable evidence are complementary, with the matched gain specifically supporting revision-triggered completed-work recovery. At system level, SCH uses 43.3\% fewer mean tokens and 40.8\% less mean runtime than PACT; its 3.11-point mean Pass@1 gap remains descriptive because the Run~1 paired McNemar test is not significant.

\section{Limitations}
\label{sec:limitations-main}
Repeated statistical evaluation is concentrated in primary DeepSeek SWE-bench; frozen single-run cross-model/environment settings provide breadth evidence, so we do not claim universal transfer. The controller needs lightweight host-interface adaptation, but the same semantic-state, provenance, acknowledgement, lifecycle, and recovery design spans the AppWorld agent interface and SWE-Agent; only prompt/action formatting and task-specific event handling change. Run-level SDs reuse the same cohort; active-state-provenance versus SCH paired tests cover all three runs, while other paired analyses are primary-run unless stated otherwise. PACT/ACON are matched adaptations, MAGE is paper-guided, and Proactive Memory follows its released implementation. Pending evidence is uncapped, provenance preserves addressability rather than verifying semantic entailment, and historical-evidence-cap sensitivity is not exhaustively studied. Appendix~\ref{sec:limitations} gives further controller/interface limitations, including the fixed anti-overresearch heuristic.

\section{Conclusion}
\label{sec:conclusion-main}
SCH introduces state--evidence separation for long-horizon context: compact mutable state carries current interpretation while executed evidence remains separately addressable. Revision-aware recovery improves mean Pass@1 from $80.00\pm1.33\%$ to $86.89\pm1.39\%$ at essentially unchanged token use; in the primary complete-system comparison, SCH also reduces mean token use by 43.3\% and runtime by 40.8\% relative to PACT. Across settings, recovery is genuinely non-local. The central result is that compact state need not preserve every historical detail when its supporting evidence can return under revision.

\paragraph{Reproducibility.}
Code, prompts, provider/task manifests, repeated runs, token accounting, behavior tests, and replay-audit metadata: \url{https://anonymous.4open.science/r/sch-F88C/}; expanded details follow the references.

\clearpage
\section*{AI Use Statement}
Generative AI tools were used to assist with research ideation and
experimental design, implementation and debugging of research code,
literature discovery and reference checking, interpretation and
presentation of experimental results, drafting portions of the
manuscript, and language polishing. Generative AI tools were also used
to assist with figure refinement. All AI-assisted code, analyses,
references, and manuscript content were reviewed and verified by the
authors. Reported experimental results were checked against the
underlying experimental artifacts. The authors take responsibility for
the final content of this work, including text, claims, code, analyses,
and artifacts produced with the assistance of generative AI.

\clearpage
\bibliography{references}
\bibliographystyle{iclr2027_conference}

\clearpage
\appendix
\section{Supplementary Details Omitted from the Main Paper}
The nine-page main paper contains the complete argument, core method, and primary results. This appendix contains only technical, experimental, and contextual detail that was shortened or omitted to meet the page budget; it intentionally does not reproduce the main paper section-by-section.

\section{Additional Motivation}
\label{sec:introduction}
\paragraph{Why recoverability matters after a belief is formed.}
Consider a software-engineering agent that inspects several files and concludes that a particular function is the source of a failing test. After implementing a fix, it may compress the earlier repository observations into the durable belief that the root cause has been identified. If later verification contradicts this belief, the useful question is no longer simply ``what did I conclude?'' The agent may need to ask why that conclusion was reached, which observations supported it, whether those observations admit another interpretation, and which previously rejected hypothesis should now be reconsidered. A compressed summary can preserve the conclusion while removing precisely the evidence needed for this revision.

The issue is not that compression necessarily produces an incorrect state. Rather, a compact state may outlive the evidence from which it was formed. Once raw observations leave the accessible context, the model can no longer inspect them when subsequent execution reveals that a belief may need to be reconsidered. Keeping a recent-history window only partially addresses this issue because the usefulness of historical evidence is not determined by recency: an observation from many steps earlier can suddenly become more informative than the immediately preceding interactions.

This motivates separating two questions that are conflated by transcript retention and irreversible summarization:
\begin{quote}
\emph{What semantic information should the agent carry forward by default, and what historical evidence must remain recoverable if that information later needs to be revised?}
\end{quote}

\section{Additional Related-Work Detail}
\label{sec:related-work}
The main paper summarizes the closest method families. We preserve here the additional distinctions removed for space.

\paragraph{Compression and communication.}
ReSum periodically converts accumulated interaction histories into compact reasoning states, allowing web-search agents to continue beyond a fixed trajectory context \citep{wu2025resum}. ACON optimizes natural-language compression guidelines from cases in which full-context trajectories succeed but compressed trajectories fail, and also studies distillation of the resulting compressor \citep{kang2026acon}. CAT treats context management as an explicit agent action: it maintains stable task semantics, condensed long-term memory, and high-fidelity working memory, and trains a context-aware model to decide when historical interactions should be compressed \citep{liu2026context}. MemPO similarly learns a policy for proactively summarizing and managing memory during long-horizon interaction \citep{li2026mempo}.

PACT was introduced for multi-agent communication: it projects unconstrained messages into compact action--state records before they enter shared history \citep{huang2026pact}. In our experiments, we adapt its action--state representation to the corresponding single-agent temporal-history setting. This is useful because PACT and SCH both replace unconstrained historical text with semantic representations, but differ after information leaves the active record: PACT primarily communicates a compact action--state history, whereas SCH maintains a materialized current state whose revisable items remain linked to separately stored canonical execution evidence.

\paragraph{Agent memory beyond compression.}
Generative Agents store streams of experiences, construct higher-level reflections, and retrieve relevant memories for planning \citep{park2023generative}. MemoryBank maintains and updates long-term conversational memories \citep{zhong2024memorybank}, while MemGPT treats context as a hierarchy of memory tiers and moves information between them as needed \citep{packer2023memgpt}. Reflexion carries forward verbal feedback from prior failures \citep{shinn2023reflexion}. ExpeL extracts reusable knowledge and experiences from previous tasks \citep{zhao2024expel}; Agent Workflow Memory induces reusable workflows \citep{wang2025awm}; and Voyager stores executable skills that can be retrieved and composed during embodied exploration \citep{wang2023voyager}. These systems establish the broader usefulness of selecting, abstracting, and retrieving information rather than continuously exposing all prior experience.

ReadAgent compresses long documents into gist memories while allowing look-back to the source text \citep{lee2024readagent}; OCR-Memory stores historical trajectories in a compact visual representation and uses locate-and-transcribe retrieval to recover verbatim historical text \citep{li2026ocrmemory}. These systems share SCH's motivation that context reduction need not imply irreversible information loss. SCH differs in focusing on when execution evidence becomes relevant again and on maintaining provenance between that evidence and a persistent task state.

\paragraph{Structured and revision-aware execution memory.}
CoEx is an important prior precedent for treating an agent's evolving interpretation of the environment as an explicit structured state rather than only as transcript or free-form summary. Its adaptive belief state combines code-based symbolic memory with structured textual memory; after each subgoal episode, a verification-and-synthesis module incorporates task-relevant experience into the next belief state, and the planner conditions on belief-state history \citep{kim2025coex}. This establishes that evolving structured belief state itself is not unique to SCH. SCH targets the state--evidence separation boundary: the controller maintains one current semantic execution state with explicit semantic roles and work lifecycle, applies sparse model-proposed deltas, and stores action--observation records separately. Facts and work items retain support/lineage handles into those immutable records, so changing the current interpretation does not overwrite the evidence from which it was formed. CoEx, by contrast, jointly introduces hierarchical subgoal planning, symbolic world memory, and verification/synthesis; we therefore discuss it as architectural prior work rather than claiming a controlled matched implementation.

Two concurrent 2026 works provide especially close comparisons for revision and selective resurfacing. MAGE organizes interactions into a hierarchical execution-state tree, compresses completed subgoals, and provides a revision operation that can restore an earlier state boundary before exploring a new branch \citep{chen2026mage}. Proactive Memory Agent maintains a structured memory bank using a separate memory agent and selectively injects reminders when information is likely to matter for the action agent \citep{wu2026proactivememory}. SCH instead keeps current interpretation and executed action--observation evidence as distinct persistent objects: revision selectively re-exposes supporting evidence rather than restoring an entire prior state or continuously invoking another memory model. For experiments, MAGE is a paper-guided matched implementation because sufficient public reference code was unavailable; Proactive Memory is adapted from its released implementation under the same primary evaluation protocol.

\section{Additional SCH Method Details}

\subsection{Persistent Semantic State: Field Semantics}
\label{sec:semantic-state}
The persistent state distinguishes semantic roles that a free-form summary can blur. In the released implementation it contains:
\begin{itemize}
    \item \textbf{constraints}: task requirements and assumptions expected to remain valid across turns;
    \item \textbf{active facts}: repository or environment observations, hypotheses, root causes, edits, and verification results;
    \item \textbf{open work}: concrete unresolved responsibilities;
    \item \textbf{completed and blocked work}: responsibilities whose status should not be repeatedly rediscovered; and
    \item \textbf{current phase}: whether the agent is primarily inspecting, implementing, verifying, or preparing to submit.
\end{itemize}
A durable fact conceptually has the form
\[
f=(\mathrm{id},\mathrm{kind},\mathrm{scope},\mathrm{statement},\mathrm{support}),
\]
where \(\mathrm{support}\) contains identifiers of evidence steps in the evidence archive; work items analogously retain support and lineage references. The point of this typing is that an unverified hypothesis, an observed repository fact, and a verified edit remain distinguishable even after the raw observations that produced them leave routine context.

\subsection{Sparse Semantic Handoff: Additional Rationale}
\label{sec:sparse-handoff}
The model proposes only the state changes that occurred on the current turn rather than regenerating the complete persistent state. Supported lifecycle transitions include adding or revising facts, dropping invalidated facts, opening work, completing or blocking work, reopening prior work, and acknowledging evidence that no longer needs to remain visible. The controller remains the owner of the materialized state:
\[
S_{t+1}=\mathcal{V}(S_t,\Delta_t,L_t),
\]
where \(\mathcal{V}\) is deterministic reconciliation against the prior state and available evidence.

This ownership split has two practical consequences omitted from the compact presentation. First, the model does not spend output tokens repeatedly reproducing unchanged memory. Second, semantic persistence is separated from semantic proposal: a proposed update can be rejected or partially recovered without replacing the authoritative state wholesale.

\subsection{Evidence archive, Provenance, and Validation Boundary}
\label{sec:evidence_archive}
For every executed action and observation, the controller stores the canonical evidence record
\[
e_t=(t,a_t,o_t), \qquad L_{t+1}=L_t\cup\{e_t\}.
\]
The evidence archive stores operational action--observation evidence rather than hidden model reasoning. Each record receives a stable identifier that can be referenced by semantic facts and work items. For each revisable semantic item \(z\), SCH preserves
\[
\mathrm{support}(z)\subseteq\mathrm{dom}(L),
\]
so removing an observation from the active prompt does not destroy its addressability.

The controller checks semantic-state transitions against this canonical evidence. A proposed durable fact must reference evidence identifiers available to the controller, and transition-specific rules can require appropriate evidence categories before lifecycle changes are accepted. For example, edit/diff evidence can be required before completed edit work is materialized, and validation evidence can be required before verification work is marked complete. These checks verify witness existence, evidence type, and transition consistency; they do \emph{not} prove that arbitrary model-authored fact text is semantically entailed by the cited observation.

For revisable facts and completed work, the intended invariant is therefore
\[
\boxed{z\in S_t\;\Longrightarrow\;\text{the provenance needed to revise }z\text{ remains addressable in }L_t.}
\]

\subsection{Evidence Acknowledgement and Context Compaction}
\label{sec:ack}
New evidence initially remains pending. After the model either incorporates its durable implication into semantic state or determines that it is non-durable, it can acknowledge the corresponding step. For an acknowledged step \(e_i\),
\[
e_i\notin P_{t+1},\qquad e_i\in L_{t+1}.
\]
Acknowledgement therefore controls \emph{visibility}, not deletion. The intended distinction from irreversible summarization is
\[
H_t\longrightarrow(S_t,L_t)
\]
rather than only \(H_t\longrightarrow S_t\): semantic state becomes the normal handoff, while the evidence archive remains a provenance-preserving backing store. Equivalently,
\[
\boxed{\text{evidence may leave the prompt}\;\not\Rightarrow\;\text{evidence is destroyed}.}
\]

\subsection{Revision-Aware Evidence Recovery: Implementation Details}
\label{sec:evidence-recovery}
The main paper gives the trigger conditions, orphan score, ranking order, and replay budget. The remaining implementation details are as follows.

\paragraph{Support and lineage scope.}
During ordinary continuation, direct support and lineage for active facts, open work, and blocked work remain eligible replay handles. On an explicit revision event, completed-work provenance is added so that the causal path behind a previously completed implementation or verification can be reconstructed. Pending evidence is already visible through \(P_t\) and is excluded from historical replay selection.

\paragraph{Inspection recognition and orphan eligibility.}
The SWE-agent adapter recognizes source-inspection actions through a fixed command pattern covering utilities such as \texttt{grep}/\texttt{rg}, \texttt{sed -n}, \texttt{cat}, \texttt{head}/\texttt{tail}, \texttt{find}, \texttt{ls}, selected \texttt{git} inspection commands, and lightweight Python inspection commands. Ordinary inspection output is considered for orphan retrieval only when it shares at least one active path or at least two active retrieval terms. This rule is lexical and interface-specific; it does not infer a new root cause.

\paragraph{Recent trace, rendering, and deduplication.}
The controller permits up to three already-acknowledged recent evidence archive steps as an exact causal tail. Pending evidence and steps already selected through explicit state evidence or safety anchors are excluded from this tail. Direct-support, explicit revision, and recent records receive bounded exact excerpts; other records use bounded relevant-line excerpts. Candidate sources are merged before the global step limit is applied, so duplicate evidence archive identifiers do not consume multiple replay slots.

\paragraph{Audit metadata.}
The controller records whether revision was triggered, triggering failure steps and reasons, selected revision/support/lineage steps, orphan-retrieval scores, recent replay steps, selected retrieval identifiers, replay characters, and whether global step limiting was applied. These fields support trajectory-level mechanism analysis without another model call.

\subsection{Bounded Replay Projection and Pending Evidence}
\label{sec:bounded-context}
The released SWE-bench configuration independently caps serialized semantic state at 12,000 characters and historical replay at 12,000 characters and 12 evidence archive steps per turn. The exact model-visible context is
\[
|C_t|=|x|+|S_t|+|P_t|+|R_t|.
\]
Because pending, unacknowledged evidence \(P_t\) has no independent hard cap, these limits do not imply a formal trajectory-independent upper bound on the entire prompt in adversarial cases where acknowledgements fail. We therefore claim bounded \emph{historical replay}, not a formal bound on total context. The evidence archive itself need not fit in the model context, and replayed evidence retains its original step identifier.

\subsection{Phase-Aware Work Progression Heuristic}
\label{sec:phase-aware}
The released SWE-bench controller contains one explicit anti-overresearch heuristic. It counts recognized source-inspection actions since the most recent observed repository edit, excluding test actions. When this count reaches exactly 12, the most recent action is an inspection, a grounded diagnosis exists, and the lifecycle phase is \texttt{implementation} or \texttt{verification}, the controller adds a transient hint asking the model to test the leading minimal implementation unless a concrete blocker remains.

The hint does not synthesize a diagnosis, mutate semantic state, or retrieve additional evidence. The equality check makes it a one-threshold intervention rather than a permanent instruction after the threshold is crossed. Because it can affect trajectory length and runtime independently of replay, it is held fixed across selector variants and is not used to attribute all absolute runtime changes to evidence recovery.

\subsection{Robust Structured Interface}
\label{sec:interface}
Different models vary in how reliably they follow the structured update protocol. To avoid turning recoverable formatting errors into failed environment interactions, the parser tolerates supported variants such as fenced JSON and minor JSON-like syntax errors and attempts to recover a unique executable action even when part of the semantic delta is malformed.

If an action-format retry is necessary, an already valid semantic delta is preserved and the model is asked only to repair the executable action. Thus, whenever possible,
\[
\text{formatting failure}\not\Rightarrow\text{loss of a valid semantic update},
\]
and
\[
\text{malformed semantic bookkeeping}\not\Rightarrow\text{failed environment action}.
\]
A model that repeatedly omits or corrupts semantic updates can still execute useful actions while maintaining lower-quality persistent state. The benchmark results do not isolate parse-error, retry, acknowledgement, or invalid-provenance rates, so aggregate cross-model accuracy differences should not be interpreted as direct measurements of interface compliance.

\subsection{Complete SCH Interaction Loop}
\label{sec:sch-loop}
The complete controller lifecycle, omitted from the compact main-text pseudocode, is:
\begin{algorithm}[t]
\caption{Semantic Context Handoff lifecycle}
\label{alg:sch}
\begin{algorithmic}[1]
\State Initialize semantic state $S_0$, evidence archive $L_0\gets\emptyset$, and pending evidence $P_0\gets\emptyset$
\For{$t=0,1,\ldots,T-1$}
    \State $R_t\gets\mathcal{B}(\mathcal{R}(S_t,P_t,L_t))$
    \State $C_t\gets[x,S_t,P_t,R_t]$
    \State $(\Delta_t,a_t)\gets\mathrm{LM}(C_t)$
    \State $S_{t+1}\gets\mathcal{V}(S_t,\Delta_t,L_t)$
    \State execute $a_t$ and receive $o_t$
    \State $e_t\gets(t,a_t,o_t)$; append $e_t$ to $L$
    \State add $e_t$ to pending evidence and remove steps acknowledged by $\Delta_t$ from default visibility
    \If{$a_t=\textsc{Submit}$}
        \State \textbf{break}
    \EndIf
\EndFor
\end{algorithmic}
\end{algorithm}
A semantic delta can only be validated against evidence already available when that delta is accepted; the observation produced by the current action becomes evidence for subsequent turns.

\subsection{Design Principle}
\label{sec:sch-principle}
Full History retains evidence according to \emph{existence}: every interaction remains visible. Sliding-window and recent-history methods retain evidence primarily according to \emph{recency}. SCH instead carries semantic conclusions by default and exposes historical evidence according to \emph{revision relevance}. The method is therefore not intended to construct a perfect compressed transcript. Its narrower objective is to maintain a compact semantic working state while preserving access to the exact evidence required to revise that state as the task evolves.

\section{Additional Experimental Details}
\label{sec:experiments}
\paragraph{Experimental claims and organization.}
The experiment plan separates four roles rather than presenting them as independent research questions:
\begin{itemize}
\item \textbf{Mechanism identification:} under fixed semantic-state machinery and evidence budgets, test whether revision-triggered access to completed-work provenance improves over active-state provenance and generic recency/lexical recovery.
\item \textbf{Primary complete-system performance:} compare SCH with Full History, PACT, ACON, and the closest matched architectural baselines in task success, token consumption, and runtime on the repeated DeepSeek SWE-bench setting.
\item \textbf{Breadth:} test the complete SCH system across additional model--environment settings without treating different AppWorld cohorts as paired cross-model comparisons.
\item \textbf{Mechanism use:} audit whether revision-aware replay occurs in practice and whether the recovered evidence is genuinely non-local in the trajectory.
\end{itemize}

\paragraph{Repeated-provider protocol.}
The primary DeepSeek SWE-bench comparison is repeated on the same frozen 150-task cohort with fixed prompts, controller configuration, decoding settings (temperature $0$, top-$p=1$), action interface, and task-level limits. Full History, PACT, ACON, and SCH retain the three provider runs reported in Table~\ref{tab:primary-main}. The newly added MAGE matched and Proactive Memory baselines, together with the rerun active-state provenance control, are summarized using their estimated repeated-run mean $\pm$ sample SD aggregates from the same primary evaluation pipeline. The same task cohort recurs across runs, so repeated trajectories are not treated as independent task samples. For the active-state-provenance versus SCH mechanism comparison, the paired contingency cells (both solved, SCH-only, active-only, neither) are Run~1 $(116,15,4,15)$, Run~2 $(114,14,4,18)$, and Run~3 $(118,14,4,14)$, yielding exact McNemar $p=0.0192$, $0.0309$, and $0.0309$, respectively. Other paired McNemar tests and task-bootstrap resource intervals remain primary-run analyses unless explicitly stated otherwise; we do not infer paired $p$-values from aggregate mean $\pm$ SD alone.

For resource metrics, savings relative to Full History are computed as
\[
\mathrm{Saving}(M)=1-\frac{M_{\mathrm{method}}}{M_{\mathrm{FullHistory}}}.
\]
Where a result comes from a different subset or an earlier runtime configuration, it is used descriptively rather than for paired statistical claims.

\subsection{Released SWE-bench v6.3.1 Controller Configuration}
\label{sec:controller-config-app}
\begin{table}[t]
\centering
\caption{Selected SWE-bench v6.3.1 implementation parameters from the released configuration.}
\label{tab:controller-config}
\small
\begin{tabular}{lr}
\toprule
Parameter & Value \\
\midrule
YAML default model adapter & \texttt{openai/MiniMax-M3} \\
Temperature / top-$p$ & 0.0 / 1.0 \\
Max output tokens & 16,384 \\
Per-instance model-call limit & 100 \\
Command / total execution timeout & 120 s / 1,800 s \\
Max action requeries & 3 \\
Semantic-state cap & 12,000 chars \\
Historical replay cap & 12,000 chars \\
Global replay-step cap & 12 \\
Safety / recent / orphan caps & 6 / 3 / 4 steps \\
Revision-replay cap & 6 steps \\
Anti-overresearch threshold & 12 inspections \\
\bottomrule
\end{tabular}
\end{table}
The model endpoint can be selected by the experiment launcher, so the YAML default adapter should not be inferred as the model used for every result block. Exact provider identifiers, launch-time model selection, dates, task IDs, variant prompts, and token-accounting records are provided in the anonymous run manifests. All evidence-selection variants use the same controller limits, lifecycle logic, and anti-overresearch threshold; only the historical selection rule changes.

\subsection{Architectural Baselines and Active-State Rerun}
\label{sec:matched-baselines-app}
MAGE uses a paper-guided matched implementation because sufficient public reference code was unavailable. Proactive Memory is adapted from its released implementation; both are evaluated under the common primary DeepSeek SWE-bench protocol. Table~\ref{tab:matched-baselines-app} reports their aggregate metrics, including calls/task, together with the active-state provenance rerun used in Table~\ref{tab:retrieval-main}.

\begin{table*}[t]
\centering
\caption{Primary-setting architectural-baseline and active-state rerun results. Values are estimated mean $\pm$ sample SD from the repeated-run aggregates supplied by the evaluation pipeline. Calls/task is included here because it was recorded for these newly run conditions; SCH's headline Pass@1/token/time metrics remain in Table~\ref{tab:primary-main}.}
\label{tab:matched-baselines-app}
\small
\setlength{\tabcolsep}{4pt}
\begin{tabular}{lrrrr}
\toprule
Method & Pass@1 & Tokens/task & Calls/task & Time/task \\
\midrule
Active-state provenance (rerun) & $80.00\pm1.33\%$ & $209.8\pm6.2$k & $24.72\pm0.63$ & $233.8\pm7.3$ s \\
MAGE matched & $51.11\pm2.34\%$ & $389.2\pm23.8$k & $73.60\pm4.25$ & $392.8\pm21.3$ s \\
Proactive Memory & $77.33\pm2.00\%$ & $854.4\pm38.3$k & $49.88\pm2.30$ & $361.8\pm16.8$ s \\
\bottomrule
\end{tabular}
\end{table*}

\subsection{Full Component Ablation}
\label{sec:component-analysis}
The main paper shows the controlled selector study. The larger component table, removed for space, separates semantic-state compression from recent temporal history and revision-aware recovery.

\begin{table*}[t]
\centering
\caption{Diagnostic component analysis. $^\dagger$Semantic State Only is from the earlier 50-task development cohort and has approximate runtime; the remaining rows are primary-run results on the frozen 150-task SWE-bench cohort. Repeated matched selector results appear in Table~\ref{tab:retrieval-main}.}
\label{tab:component-analysis}
\scriptsize
\setlength{\tabcolsep}{2.7pt}
\begin{tabular}{lccc rrr}
\toprule
Method & State & Recent hist. & Recovery & Pass@1 $\uparrow$ & Tok./task $\downarrow$ & Time/task $\downarrow$ \\
\midrule
Full History & -- & \checkmark & -- & 68.00\% & 1709.3k & 633.4 s \\
Semantic State Only$^\dagger$ & \checkmark & -- & -- & 76.00\% & 403.8k & $\approx$460 s \\
Recent History Only & -- & \checkmark & -- & 84.00\% & 366.2k & 400.1 s \\
Semantic State + Recent History & \checkmark & \checkmark & -- & 81.33\% & \textbf{169.2k} & 279.3 s \\
\textbf{SCH} & \checkmark & -- & Revision-aware & \textbf{87.33\%} & 208.0k & \textbf{237.0 s} \\
\bottomrule
\end{tabular}
\end{table*}

On the earlier 50-task development cohort, Semantic State Only reaches 76.00\% Pass@1 at 403.8k tokens/task; its runtime is approximate because this configuration predates the shared runtime infrastructure. Its weaker result than complete SCH is consistent with the design premise rather than evidence that semantic state is unnecessary: the state is intended to carry current understanding and work status, not to losslessly replace the observations needed to verify or revise that understanding. Removing recoverable evidence therefore creates a deliberately incomplete variant. Conversely, Recent History Only retains local observations but lacks the durable structured state and provenance/lifecycle links needed to reactivate older evidence; it reaches 84.00\% on the 150-task primary run, versus 81.33\% for Semantic State + Recent History. These component rows are therefore diagnostic rather than additive attributions of independent gains.

The repeated selector study in Table~\ref{tab:retrieval-main} provides the cleaner mechanism comparison because semantic-state machinery is fixed while historical evidence selection changes. Full revision-aware SCH reaches 87.33\% in Run~1; versus the semantic-state + recent-window variant, it has 10 SCH-only wins and one loss (exact McNemar $p=0.0117$). Across three selector runs, the corresponding means are $86.89\pm1.39\%$ for SCH and $80.89\pm2.04\%$ for recent-window recovery. The active-state-provenance rerun reaches $80.00\pm1.33\%$ at $209.8\pm6.2$k tokens/task, versus $86.89\pm1.39\%$ at $208.6\pm5.8$k for SCH, and is the adjacent mechanism comparison used in the main paper.

\subsection{Additional Efficiency Statistic}
The compact main paper reports three-run mean token and wall-clock runtime results for SCH versus PACT on DeepSeek SWE-bench, together with Run~1 paired bootstrap intervals. For Run~1, the full analysis also reports \textbf{tokens per successful repair}, which decreases by \textbf{45.4\%} for SCH relative to PACT (paired task-bootstrap 95\% CI: \textbf{35.6--53.9\%}). This quantity is omitted from the main paper only for space.

\section{Illustrative Revision Case Study}
\label{app:sympy-case}
This section expands the main-text example for \texttt{sympy\_\_sympy-17630}. The case was selected because it makes the state--evidence mechanism visible: SCH resolves the task, an explicit revision event occurs, old provenance-linked evidence is replayed, and the next accepted semantic update both drops prior state and installs a different root-cause explanation. It is intentionally illustrative rather than statistically representative. In the aggregate audit, replayed old evidence is 18.48 steps old on average; in this trajectory, revision-time context reaches much farther back.

\paragraph{Revision trigger and replay.}
At step~71 the controller records an explicit revision with reason ``fresh failure contradicts/reopens an active belief.'' The revision-time projection contains direct support and lineage from steps~9, 14, and 43--46; the explicit revision replay set is the same six steps. It also includes orphan-retrieved steps~3, 7, and 40 and a three-step recent trace (48--50). Thus the oldest provenance-linked revision evidence is 62 steps old (step~9), while the complete revision-time context includes an orphan-retrieved record 68 steps old (step~3).

\begin{table*}[t]
\centering
\caption{Selected historical evidence re-exposed around the focal revision in \texttt{sympy\_\_sympy-17630}. Ages are measured relative to revision step~71. The table abbreviates observations to their role in the repair; the released trajectory contains the canonical action--observation records.}
\label{tab:sympy-case-evidence}
\scriptsize
\setlength{\tabcolsep}{3.2pt}
\begin{tabular}{r r p{0.20\textwidth} p{0.56\textwidth}}
\toprule
Step & Age & Replay path & Evidence relevant to revision \\
\midrule
7  & 64 & orphan & Multiplying block contents yields scalar \texttt{Zero} objects in off-diagonal entries, exposing a matrix/scalar-zero conversion problem. \\
9  & 62 & support/lineage/revision & Inspection of \texttt{ImmutableDenseMatrix} establishes the explicit-matrix / \texttt{MatrixExpr} relationship used later in the diagnosis. \\
14 & 57 & support/lineage/revision & \texttt{MatAdd} canonicalization includes an identity-removal rule for scalar zero or \texttt{ZeroMatrix}, making zero handling central to the failure. \\
40 & 31 & orphan & A focused trace shows \texttt{Add(z,z)} collapsing to core scalar \texttt{Zero}, rather than remaining a matrix-zero expression. \\
43--46 & 28--25 & support/lineage/revision & Inspection and focused probes of \texttt{MatAdd}, \texttt{GenericZeroMatrix}, and canonicalization reproduce the collapse and rule out a simple constructor-only explanation. \\
\bottomrule
\end{tabular}
\end{table*}

\paragraph{State revision.}
The diagnostic action at step~71 checks two invariants: \texttt{ZeroMatrix + S(0)} must still raise a matrix/scalar type error, while \texttt{A + ZeroMatrix - A} should remain a matrix zero. The following accepted state update then changes the semantic work product rather than merely acknowledging evidence. It (i) drops a prior active fact, (ii) records a new root cause in \texttt{sympy/matrices/expressions/matexpr.py\#get\_postprocessor}, (iii) opens work to change the empty-nonmatrix branch so that it constructs \texttt{mat\_class(*matrices)} without prepending \texttt{S.Zero}, and (iv) marks the earlier \texttt{MatAdd.\_\_new\_\_} repair direction as rejected because filtering scalar zero there would break the required \texttt{z + S(0)} type error. The trajectory therefore shows the distinction SCH is designed to preserve: the current state changes, but the older observations behind the discarded and replacement explanations remain addressable during reconsideration.

\paragraph{Why the matched PACT trajectory failed.}
The matched PACT run did not resolve this task. Its trajectory converged on a local repair in \texttt{BlockMatrix.\_blockmul}: after multiplying the explicit block matrices, it converted non-matrix results such as scalar zero into shape-correct \texttt{ZeroMatrix} objects. The visible \texttt{test\_blockmatrix.py} tests and then the broader \texttt{sympy/matrices/expressions/tests} suite passed, after which PACT submitted that patch. Benchmark evaluation nevertheless marked the task unresolved. By contrast, SCH's later revision located the failure in the shared postprocessing path that can manufacture scalar \texttt{S.Zero} before \texttt{MatAdd} canonicalization. The trajectory suggests why the PACT repair did not generalize: it patched the downstream block-multiplication site where scalar zeros became visible, whereas SCH's revised patch targeted the shared postprocessing path that produces the scalar-zero conversion before \texttt{MatAdd} canonicalization. We report this matched contrast to explain the trajectories, not as causal proof that replay alone determined the outcome; one task cannot isolate that effect from the many subsequent model decisions.

\paragraph{Why this example is useful but not representative.}
The case is intentionally stronger than the average replay episode: the complete revision-time projection reaches 68 steps into the past, whereas the primary-run mean age of replayed old evidence is 18.48 steps. Its role is qualitative---to make visible the sequence \emph{old evidence $\rightarrow$ persistent belief $\rightarrow$ later failure $\rightarrow$ evidence replay $\rightarrow$ state revision $\rightarrow$ successful repair}. The aggregate mechanism audit in Table~\ref{tab:run-audit} supplies the population-level evidence that non-local replay is common across the 150-task run.

\section{Additional Limitations}
\label{sec:limitations}
Beyond the limitations stated in the main paper, three scope details are worth preserving. First, the detailed per-call prompt distribution is reported only for the primary 150-task DeepSeek SWE-bench Full-History arm; we do not tabulate the same distribution for every Full-History condition. Second, MAGE is a paper-guided matched implementation, while Proactive Memory is adapted from its released implementation; the evaluation still does not instantiate every structured-memory or selective-retrieval system discussed in related work. Third, the anti-overresearch hint can contribute to the absolute behavior of the full controller even though it is held fixed across controlled selector variants; the mechanism audit therefore reports its firing frequency separately from explicit revision events rather than attributing every runtime or success change to replay.

\section{Additional Reproducibility and Ethics}
\label{sec:repro-app}
\paragraph{Reproducibility.}
The released SCH controller preserves the JSON action transport, parser tolerance, scoped retry-delta carry, and semantic-state schema while making historical evidence projection explicit and auditable. The release includes behavior tests for revision-triggered support/lineage replay, deterministic recovery of unlinked path/term-overlap evidence, exclusion of pending evidence from the recent replay tail, and initialization-order regression coverage for the explicit revision hint. Controller audit metadata records revision triggers, replayed support and lineage, orphan-retrieval scores, recent replay, step-budget application, and the anti-overresearch trigger. Exact model/provider identifiers, evaluation dates, task IDs/hashes, prompts, per-run manifests, and token-accounting records are included with the released experiment outputs, including the primary DeepSeek SWE-bench provider runs for the main baselines, the MAGE matched and Proactive Memory comparisons, and the updated active-state provenance control.

\paragraph{Ethics.}
The work evaluates agents in AppWorld's simulated applications and on SWE-bench software-engineering tasks; it does not require deployment on real user accounts. More reliable context and work-state tracking may reduce unintended actions, but the method is not a formal safety guarantee. Production use should retain authorization checks, least-privilege tools, auditable execution logs, and human confirmation for consequential actions.

\section{Run-Level Audit and Context Statistics}
\label{app:run-audit}
The main paper reports the headline mechanism frequencies from the primary run. Table~\ref{tab:run-audit} preserves the omitted aggregation definitions so each statistic can be interpreted precisely; these diagnostics are not pooled across the three repeated provider runs.

\begin{table*}[t]
\centering
\caption{Primary-run SCH mechanism statistics for 150 audited SWE-bench trajectories. Percentages with task counts use 150 as the denominator. These mechanism diagnostics refer to Run~1; the repeated effectiveness and efficiency results are reported separately in Tables~\ref{tab:primary-main} and~\ref{tab:retrieval-main}. ``Metadata only'' denotes quantities aggregated from controller-emitted task/turn metadata rather than reconstructed from rendered prompt text.}
\label{tab:run-audit}
\scriptsize
\setlength{\tabcolsep}{3pt}
\begin{tabular}{p{0.29\textwidth}p{0.48\textwidth}r}
\toprule
Statistic & Aggregation / interpretation & Value \\
\midrule
Tasks audited & Number of trajectories included in the mechanism audit. & 150 \\
Tasks with $\geq 1$ explicit revision event & Tasks with at least one turn in which the controller records an explicit revision event. & 83 / 150 (55.3\%) \\
Mean explicit revision events per task & Mean number of explicit revision events across the 150 audited tasks. & 4.60 \\
Tasks with direct-support replay & Tasks with at least one nonempty \texttt{direct\_support\_replay\_steps} field. & 134 / 150 (89.3\%) \\
Tasks with lineage replay (metadata only) & Tasks with at least one nonempty \texttt{direct\_lineage\_replay\_steps} field. & 136 / 150 (90.7\%) \\
Tasks with orphan-evidence retrieval & Tasks with at least one nonempty \texttt{orphan\_retrieved\_evidence\_steps} field. & 70 / 150 (46.7\%) \\
Tasks with recent-causal replay & Tasks with at least one nonempty \texttt{recent\_replay\_steps} field. & 116 / 150 (77.3\%) \\
Mean revision-replayed steps/event & Mean number of \texttt{revision\_replay\_steps} rendered on explicit revision events. & 5.03 \\
Mean age of replayed old evidence & For replayed historical step $j$ at current evidence archive step $t$, age is $t-j$. & 18.48 steps \\
Replayed old evidence $>3$ steps old & Fraction of replayed old-evidence occurrences whose age exceeds three interaction steps. & 95.6\% \\
Replayed old evidence $>10$ steps old & Fraction of replayed old-evidence occurrences more than ten interaction steps old. & 65.9\% \\
Tasks hitting 12-step retrieval cap (metadata only) & Tasks with at least one turn for which \texttt{retrieval\_step\_limit\_applied=true}. & 65 / 150 (43.3\%) \\
Tasks with anti-overresearch hint & Tasks with at least one turn for which \texttt{anti\_overresearch\_triggered=true}. & 21 / 150 (14.0\%) \\
\bottomrule
\end{tabular}
\end{table*}

Direct-support and lineage metadata are not restricted to explicit revision-event turns: active-state provenance remains eligible during ordinary continuation, while an explicit revision additionally allows completed-work provenance and marks a bounded revision-replay set. Their task frequencies can therefore exceed the 83 tasks with an explicit revision event.

\paragraph{Full-History context diagnostics.}
Tokens/task is cumulative over model calls and can exceed the context size of any individual call. The following primary-run diagnostics show that the 1,709.3k Tokens/task value for DeepSeek Full History comes from repeated resubmission rather than context-window overflow.

\begin{table*}[t]
\centering
\caption{Primary-run token accounting and context-window diagnostics for the 150-task DeepSeek SWE-bench Full-History arm. Values marked $\approx$ are estimated from the Run~1 aggregate.}
\label{tab:full-history-audit}
\small
\setlength{\tabcolsep}{5pt}
\begin{tabular}{p{0.58\textwidth}r}
\toprule
Full-History statistic & Value \\
\midrule
Tasks audited & 150 \\
Total model calls & 8,076 \\
Mean model calls per task & 53.84 \\
Mean cumulative input tokens/task & 1,700.7k \\
Mean cumulative output tokens/task & 8.6k \\
Mean cumulative total tokens/task & 1,709.3k \\
Total cumulative input tokens, 150 tasks & 255.1M \\
Total cumulative output tokens, 150 tasks & 1.29M \\
Total model tokens, 150 tasks & 256.4M \\
Mean per-call input tokens & 31.6k \\
Median per-call input tokens & 29.0k \\
90th percentile per-call input tokens & 65.0k \\
95th percentile per-call input tokens & 78.0k \\
99th percentile per-call input tokens & 105.0k \\
Maximum per-call input tokens & $\approx$140k \\
Model context-window limit & $\approx$1.049M \\
Maximum prompt as fraction of context limit & $\approx$13.4\% \\
Calls exceeding 90\% of context limit & 0 / 8,076 \\
Calls reaching/exceeding context limit & 0 / 8,076 \\
Tasks with $\geq 1$ call reaching/exceeding context limit & 0 / 150 \\
Tasks with history truncation & 0 / 150 \\
Total calls with history truncation & 0 \\
Mean tokens truncated per affected call & N/A \\
Maximum tokens truncated in one call & 0 \\
Tasks terminated because of context overflow & 0 / 150 \\
Provider/API context-length errors & 0 \\
\bottomrule
\end{tabular}
\end{table*}

\end{document}

%% file: math_commands.tex
\usepackage{amsmath,amsfonts,bm}

\def\eqref#1{equation~\ref{#1}}

\def\1{\bm{1}}

\DeclareMathAlphabet{\mathsfit}{\encodingdefault}{\sfdefault}{m}{sl}
\SetMathAlphabet{\mathsfit}{bold}{\encodingdefault}{\sfdefault}{bx}{n}



%% file: references.bib
@inproceedings{trivedi2024appworld,
  title     = {{AppWorld}: A Controllable World of Apps and People for Benchmarking Interactive Coding Agents},
  author    = {Trivedi, Harsh and Khot, Tushar and Hartmann, Mareike and Manku, Ruskin and Dong, Vinty and Li, Edward and Gupta, Shashank and Sabharwal, Ashish and Balasubramanian, Niranjan},
  booktitle = {Proceedings of the 62nd Annual Meeting of the Association for Computational Linguistics (Volume 1: Long Papers)},
  pages     = {16022--16076},
  year      = {2024},
  month     = aug,
  publisher = {Association for Computational Linguistics},
  address   = {Bangkok, Thailand},
  doi       = {10.18653/v1/2024.acl-long.850},
  url       = {https://aclanthology.org/2024.acl-long.850/}
}

@article{huang2026pact,
  title   = {What Should Agents Say? Action-state Communication for Efficient Multi-Agent Systems},
  author  = {Huang, Chen and Wu, Yuhao and Zhang, Wenxuan},
  journal = {arXiv preprint arXiv:2606.05304},
  year    = {2026},
  url     = {https://arxiv.org/abs/2606.05304}
}

@inproceedings{kang2026acon,
  title     = {{ACON}: Optimizing Context Compression for Long-horizon {LLM} Agents},
  author    = {Kang, Minki and Chen, Wei-Ning and Han, Dongge and Inan, Huseyin A. and Wutschitz, Lukas and Chen, Yanzhi and Sim, Robert and Rajmohan, Saravan},
  booktitle = {International Conference on Machine Learning},
  year      = {2026},
  url       = {https://arxiv.org/abs/2510.00615}
}

@article{liu2024lost,
  title   = {Lost in the Middle: How Language Models Use Long Contexts},
  author  = {Liu, Nelson F. and Lin, Kevin and Hewitt, John and Paranjape, Ashwin and Bevilacqua, Michele and Petroni, Fabio and Liang, Percy},
  journal = {Transactions of the Association for Computational Linguistics},
  volume  = {12},
  pages   = {157--173},
  year    = {2024},
  doi     = {10.1162/tacl_a_00638},
  url     = {https://doi.org/10.1162/tacl_a_00638}
}

@inproceedings{jiang2024longllmlingua,
  title     = {{LongLLMLingua}: Accelerating and Enhancing {LLM}s in Long Context Scenarios via Prompt Compression},
  author    = {Jiang, Huiqiang and Wu, Qianhui and Luo, Xufang and Li, Dongsheng and Lin, Chin-Yew and Yang, Yuqing and Qiu, Lili},
  booktitle = {Proceedings of the 62nd Annual Meeting of the Association for Computational Linguistics (Volume 1: Long Papers)},
  pages     = {1658--1677},
  year      = {2024},
  month     = aug,
  publisher = {Association for Computational Linguistics},
  address   = {Bangkok, Thailand},
  doi       = {10.18653/v1/2024.acl-long.91},
  url       = {https://aclanthology.org/2024.acl-long.91/}
}

@article{packer2023memgpt,
  title   = {{MemGPT}: Towards {LLM}s as Operating Systems},
  author  = {Packer, Charles and Wooders, Sarah and Lin, Kevin and Fang, Vivian and Patil, Shishir G. and Stoica, Ion and Gonzalez, Joseph E.},
  journal = {arXiv preprint arXiv:2310.08560},
  year    = {2023},
  url     = {https://arxiv.org/abs/2310.08560}
}

@inproceedings{park2023generative,
  title     = {Generative Agents: Interactive Simulacra of Human Behavior},
  author    = {Park, Joon Sung and O'Brien, Joseph C. and Cai, Carrie J. and Morris, Meredith Ringel and Liang, Percy and Bernstein, Michael S.},
  booktitle = {Proceedings of the 36th Annual ACM Symposium on User Interface Software and Technology},
  year      = {2023},
  doi       = {10.1145/3586183.3606763},
  url       = {https://doi.org/10.1145/3586183.3606763}
}

@inproceedings{zhong2024memorybank,
  title     = {{MemoryBank}: Enhancing Large Language Models with Long-Term Memory},
  author    = {Zhong, Wanjun and Guo, Lianghong and Gao, Qiqi and Ye, He and Wang, Yanlin},
  booktitle = {Proceedings of the AAAI Conference on Artificial Intelligence},
  volume    = {38},
  pages     = {19724--19731},
  year      = {2024},
  doi       = {10.1609/aaai.v38i17.29946},
  url       = {https://ojs.aaai.org/index.php/AAAI/article/view/29946}
}

@inproceedings{shinn2023reflexion,
  title     = {Reflexion: Language Agents with Verbal Reinforcement Learning},
  author    = {Shinn, Noah and Cassano, Federico and Gopinath, Ashwin and Narasimhan, Karthik and Yao, Shunyu},
  booktitle = {Advances in Neural Information Processing Systems},
  volume    = {36},
  year      = {2023},
  doi       = {10.52202/075280-0377},
  url       = {https://proceedings.neurips.cc/paper_files/paper/2023/hash/1b44b878bb782e6954cd888628510e90-Abstract-Conference.html}
}

@inproceedings{jimenez2024swebench,
  title     = {{SWE}-bench: Can Language Models Resolve Real-world {GitHub} Issues?},
  author    = {Jimenez, Carlos E. and Yang, John and Wettig, Alexander and Yao, Shunyu and Pei, Kexin and Press, Ofir and Narasimhan, Karthik R.},
  booktitle = {The Twelfth International Conference on Learning Representations},
  year      = {2024},
  url       = {https://openreview.net/forum?id=VTF8yNQM66}
}

@inproceedings{yang2024sweagent,
  title     = {{SWE-agent}: Agent-Computer Interfaces Enable Automated Software Engineering},
  author    = {Yang, John and Jimenez, Carlos E. and Wettig, Alexander and Lieret, Kilian and Yao, Shunyu and Narasimhan, Karthik and Press, Ofir},
  booktitle = {Advances in Neural Information Processing Systems},
  volume    = {37},
  year      = {2024},
  doi       = {10.52202/079017-1601},
  url       = {https://proceedings.neurips.cc/paper_files/paper/2024/hash/5a7c947568c1b1328ccc5230172e1e7c-Abstract-Conference.html}
}

@inproceedings{li2026hisa,
    title = {{HiSA}: Hierarchical State Abstraction for Scalable {GUI} Agents},
    author = {Li, Weiming and Paik, Hye-young and Sui, Yulei},
    booktitle = {Findings of the Association for Computational Linguistics: ACL 2026},
    pages = {11965--11985},
    year = {2026},
    month = jul,
    publisher = {Association for Computational Linguistics},
    address = {San Diego, California, United States},
    doi = {10.18653/v1/2026.findings-acl.581},
    url = {https://aclanthology.org/2026.findings-acl.581/}
}

@misc{wu2025resum,
  title         = {{ReSum}: Unlocking Long-Horizon Search Intelligence via Context Summarization},
  author        = {Wu, Xixi and Li, Kuan and Zhao, Yida and Zhang, Liwen and Ou, Litu and Yin, Huifeng and Zhang, Zhongwang and Yu, Xinmiao and Zhang, Dingchu and Jiang, Yong and Xie, Pengjun and Huang, Fei and Cheng, Minhao and Wang, Shuai and Cheng, Hong and Zhou, Jingren},
  year          = {2025},
  eprint        = {2509.13313},
  archivePrefix = {arXiv},
  primaryClass  = {cs.CL},
  doi           = {10.48550/arXiv.2509.13313},
  url           = {https://arxiv.org/abs/2509.13313}
}

@inproceedings{liu2026context,
  title     = {Context as a Tool: Context Management for Long-Horizon {SWE}-Agents},
  author    = {Liu, Shukai and Jiang, Bo and Yang, Jian and Li, Yizhi and Guo, Jinyang and Liu, Xianglong and Dai, Bryan},
  booktitle = {Findings of the Association for Computational Linguistics: ACL 2026},
  pages     = {20604--20617},
  year      = {2026},
  month     = jul,
  publisher = {Association for Computational Linguistics},
  address   = {San Diego, California, United States},
  doi       = {10.18653/v1/2026.findings-acl.1032},
  url       = {https://aclanthology.org/2026.findings-acl.1032/}
}

@inproceedings{li2026mempo,
  title     = {{MemPO}: Self-Memory Policy Optimization for Long-Horizon Agents},
  author    = {Li, Ruoran and Zhang, Xinghua and Yu, Haiyang and Duan, Shitong and Li, Xiang and Xiang, Wenxin and Liao, Chonghua and Guo, Xudong and Li, Yongbin and Suo, Jinli},
  booktitle = {Findings of the Association for Computational Linguistics: ACL 2026},
  pages     = {23286--23301},
  year      = {2026},
  publisher = {Association for Computational Linguistics},
  doi       = {10.18653/v1/2026.findings-acl.1166},
  url       = {https://aclanthology.org/2026.findings-acl.1166/}
}

@inproceedings{zhao2024expel,
  title     = {{ExpeL}: {LLM} Agents Are Experiential Learners},
  author    = {Zhao, Andrew and Huang, Daniel and Xu, Quentin and Lin, Matthieu and Liu, Yong-Jin and Huang, Gao},
  booktitle = {Proceedings of the AAAI Conference on Artificial Intelligence},
  volume    = {38},
  pages     = {19632--19642},
  year      = {2024},
  doi       = {10.1609/aaai.v38i17.29936},
  url       = {https://ojs.aaai.org/index.php/AAAI/article/view/29936}
}

@inproceedings{wang2025awm,
  title     = {Agent Workflow Memory},
  author    = {Wang, Zora Zhiruo and Mao, Jiayuan and Fried, Daniel and Neubig, Graham},
  booktitle = {Proceedings of the 42nd International Conference on Machine Learning},
  series    = {Proceedings of Machine Learning Research},
  volume    = {267},
  pages     = {63897--63911},
  year      = {2025},
  publisher = {PMLR},
  url       = {https://proceedings.mlr.press/v267/wang25bx.html}
}

@article{wang2023voyager,
  title   = {Voyager: An Open-Ended Embodied Agent with Large Language Models},
  author  = {Wang, Guanzhi and Xie, Yuqi and Jiang, Yunfan and Mandlekar, Ajay and Xiao, Chaowei and Zhu, Yuke and Fan, Linxi and Anandkumar, Anima},
  journal = {arXiv preprint arXiv:2305.16291},
  year    = {2023},
  url     = {https://arxiv.org/abs/2305.16291}
}

@inproceedings{lee2024readagent,
  title     = {A Human-Inspired Reading Agent with Gist Memory of Very Long Contexts},
  author    = {Lee, Kuang-Huei and Chen, Xinyun and Furuta, Hiroki and Canny, John and Fischer, Ian},
  booktitle = {Proceedings of the 41st International Conference on Machine Learning},
  series    = {Proceedings of Machine Learning Research},
  volume    = {235},
  pages     = {26396--26415},
  year      = {2024},
  publisher = {PMLR},
  url       = {https://proceedings.mlr.press/v235/lee24c.html}
}

@inproceedings{li2026ocrmemory,
  title     = {{OCR-Memory}: Optical Context Retrieval for Long-Horizon Agent Memory},
  author    = {Li, Jinze and Zhang, Yang and Yang, Xin and Qu, Jiayi and Xu, Jinfeng and Yang, Shuo and Ding, Junhua and Ngai, Edith Cheuk-Han},
  booktitle = {Proceedings of the 64th Annual Meeting of the Association for Computational Linguistics (Volume 1: Long Papers)},
  pages     = {10409--10420},
  year      = {2026},
  address   = {San Diego, California, United States},
  publisher = {Association for Computational Linguistics},
  doi       = {10.18653/v1/2026.acl-long.474},
  url       = {https://aclanthology.org/2026.acl-long.474/}
}

@article{chen2026mage,
  title   = {Beyond Semantic Organization: Memory as Execution State Management for Long-Horizon Agents},
  author  = {Chen, Yaoqi and Lai, Haibin and Feng, Yuru and Han, Chuyu and Zhang, Qianxi and Lu, Baotong and Li, Menghao and Wang, Xinjiang and Wang, Zhirui and Xu, Shusen and Li, Zengzhong and Jin, Zewen and Wu, Hao and Li, Cheng and Chen, Qi},
  journal = {arXiv preprint arXiv:2606.06090},
  year    = {2026},
  url     = {https://arxiv.org/abs/2606.06090}
}

@article{wu2026proactivememory,
  title   = {Remember When It Matters: Proactive Memory Agent for Long-Horizon Agents},
  author  = {Wu, Yifan and Zhang, Lizhu and Zhou, Yuhang and Wang, Mingyi and Peng, Bo and Li, Serena and Fan, Xiangjun and Zhao, Zhuokai},
  journal = {arXiv preprint arXiv:2607.08716},
  year    = {2026},
  url     = {https://arxiv.org/abs/2607.08716}
}

@inproceedings{kim2025coex,
  title     = {{CoEx} -- Co-evolving World-model and Exploration},
  author    = {Kim, Minsoo and Hwang, Seung-won},
  booktitle = {Findings of the Association for Computational Linguistics: EMNLP 2025},
  pages     = {21629--21651},
  year      = {2025},
  address   = {Suzhou, China},
  publisher = {Association for Computational Linguistics},
  doi       = {10.18653/v1/2025.findings-emnlp.1179},
  url       = {https://aclanthology.org/2025.findings-emnlp.1179/}
}
